\pdfoutput=1

\documentclass[11pt]{article}

\usepackage{acl}

\usepackage[switch]{lineno} 

\usepackage{times}
\usepackage{latexsym}

\usepackage[T1]{fontenc}

\usepackage[utf8]{inputenc}

\usepackage{microtype}

\usepackage{graphicx}
\usepackage{CJKutf8}
\usepackage{tabularx}
\usepackage{bm}
\usepackage{amsmath}
\usepackage{multirow}

\title{Techniques to Improve Neural Math Word Problem Solvers}

\author{
  Youyuan Zhang \\
  Student \\
  Computer Science Department \\
  McGill University \\
  \texttt{youyuan.zhang@mail.mcgill.ca} \\}

\begin{document}

\maketitle
\begin{abstract}
Developing automatic Math Word Problem (MWP) solvers is a challenging task that demands the ability of understanding and mathematical reasoning over the natural language. Recent neural-based approaches mainly encode the problem text using a language model and decode a mathematical expression over quantities and operators iteratively. Note the problem text of a MWP consists of a context part and a question part, a recent work finds these neural solvers may only perform shallow pattern matching between the context text and the golden expression, where question text is not well used. Meanwhile, existing decoding processes fail to enforce the mathematical laws into the design, where the representations for mathematical equivalent expressions are different. To address these two issues, we propose a new encoder-decoder architecture that fully leverages the question text and preserves step-wise commutative law. Besides generating quantity embeddings, our encoder further encodes the question text and uses it to guide the decoding process. At each step, our decoder uses Deep Sets to compute expression representations so that these embeddings are invariant under any permutation of quantities. Experiments on four established benchmarks demonstrate that our framework outperforms state-of-the-art neural MWP solvers, showing the effectiveness of our techniques. We also conduct a detailed analysis of the results to show the limitations of our approach and further discuss the potential future work\footnote{Code is available at \url{https://github.com/sophistz/Question-Aware-Deductive-MWP}.}.

\end{abstract}

\section{Introduction}
\begin{figure}[t]
    \centering
    \includegraphics[width=\columnwidth]{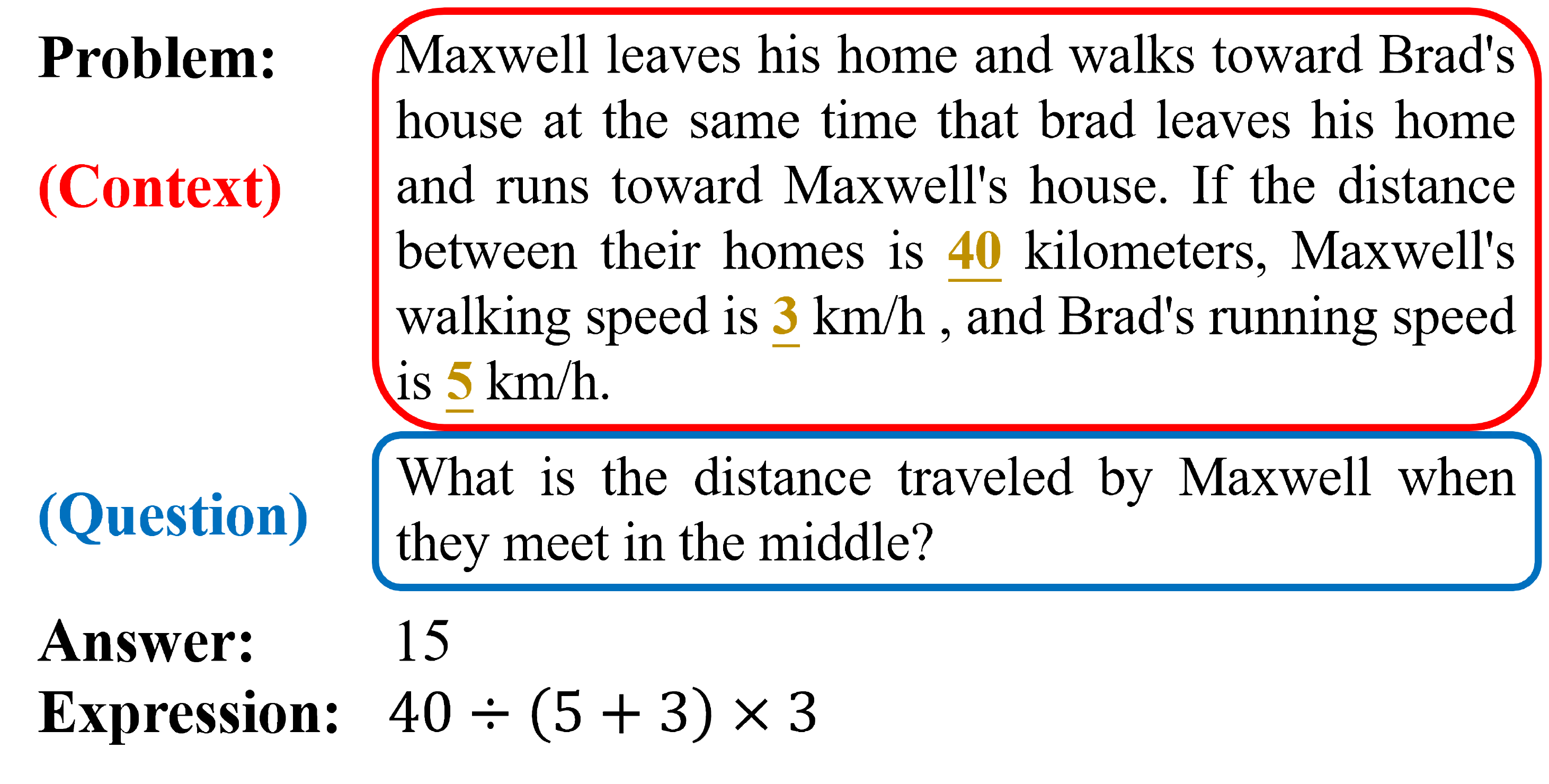}
    \caption{An example of a MWP.}
    \label{fig:1}
\end{figure}

Math word problem (MWP) solving is the task of answering a mathematical question described in natural language~\citep{bobrow1964natural}. Figure~\ref{fig:1} shows an example of a MWP. The input is composed of a context text and a question text, where the context part depicts several narratives with some quantities, and the question part defines the goal. The output is a numerical value as the answer. To obtain such an answer, one needs to infer a mathematical expression that specifies the operators over the quantities in the problem. To solve a MWP, the machine is required to not only understand the description and query of the problem but also perform mathematical reasoning over the text.

Over the last few years, with the presence of deep learning models in the field of NLP~\citep{chung2014empirical,vaswani2017attention,devlin2018bert}, a growing number of neural-based approaches are proposed for effectively solving MWPs~\citep{faldu2021towards,lan2022mwptoolkit,sundaram2022nlp}. Inspired by machine translation research~\citep{sutskever2014sequence}, a line of research~\citep{wang2017deep,amini2019mathqa} formulates the task as a sequence-to-sequence (Seq2Seq) translation problem from natural language to mathematical expression. They use an encoder to embed the problem text using a language model (e.g., GRU~\citep{chung2014empirical}, BERT~\citep{devlin2018bert}) and then employ a decoder to generate an expression sequentially. Later works~\citep{xie2019goal,zhang2020graph,cao2021bottom, jie2022learning} develop Seq2Tree or Seq2DAG neural models, where the decoders produce operators and quantities based on a specific structure iteratively. Despite the recent progress, these neural solvers are still miles away from building robust representations and effective solutions like humans. Some work~\citep{patel2021nlp} finds existing neural-based solvers that do not have access to the question part asked in the problem can still solve a large fraction of MWPs, indicating these solvers may only perform shallow reasoning or pattern matching rather than fully understanding and reasoning over the mathematical text. Additionally, most neural solvers do not preserve the mathematical laws (e.g., commutative law) when decoding an expression~\citep{faldu2021towards}, where the representations for mathematical equivalent expressions are not the same. For example, the hidden representation for the expression $(1 - 2) + 3$ is different from that of the expression $(-2 + 1) + 3$, although they are computing the same quantity.

To mitigate the two issues mentioned above, we propose a novel neural framework that fully exploits the question part of a problem and enforces step-wise commutative law. Specifically, after feeding both the context and the question into an encoder, we obtain the representation of the question text and then regard it as a query to guide the decoding process. This mimics the problem-solving procedure like humans: we often seek the answer to a problem conditioned on our goal. When generating a new expression from quantities and operators, we apply Deep Sets~\citep{zaheer2017deep} for commutative operators like addition and multiplication, and MLPs for non-commutative operators like additive inverse and multiplicative inverse. Such a design preserves commutative law at each step so that it is more robust to make predictions. To evaluate our proposed approach, we experiment on four multilingual MWP benchmarks (MAWPS~\citep{koncel2016mawps}, Math23k~\citep{wang2017deep}, SVAMPS~\citep{patel2021nlp} and MathQA~\citep{amini2019mathqa}), showing our framework outperforms state-of-the-art (SOTA) neural-based solvers. Detailed analysis of the results is also conducted to discuss the limitations and potential improvement.

Our contributions are summarized as follows:
\begin{itemize}
    \item We propose a novel encoder-decoder framework that fully leverages the question part of a MWP and enforces step-wise commutative law to improve MWP solving.
    \item Our experimental results on four established benchmarks across two languages show that our model outperforms SOTA neural-based MWP solvers. We further analyze the experimental results and discuss the limitations of our techniques and potential future work.
\end{itemize}

\section{Related Work}
Research on solving MWP has a long history back to the 1960s~\citep{bobrow1964natural}. Most early works~\citep{fletcher1985understanding, dellarosa1986computer} employ a rule-based method to convert text input to an underlying pre-defined schema. While this slot-filling-like approach is robust to handle irrelevant information, the rules are hard to exhaustively capture the myriad nuances of language and thus do not generalize well across varying language styles. 

With the development of machine learning, a stream of research~\citep{kushman2014learning,zhou2015learn,mitra2016learning,roy2018mapping} leverages statistical machine learning techniques into MWP solving. These techniques score several expressions within an optimization-based score framework and subsequently arrive at the correct mathematical model for the given text. However, similar to the rule-based methods, these methods discover the expression templates from the training data and do not generalize to the unseen templates during the inference time.

Recently, learning-based models have become a new trend in solving MWPs~\citep{faldu2021towards,lan2022mwptoolkit,sundaram2022nlp}. These neural solvers mainly first encode the problem text into the latent space and then decode the expression using operators and quantities iteratively. \citet{wang2017deep} is the seminal work that adopts recurrent neural networks as the encoder and decoder to generate target equations in a Seq2Seq fashion. Following works enhance the Seq2Seq model with reinforcement learning~\citep{huang2018neural}, multi-head attention~\citep{li2019modeling}, and large language models~\citep{tan2021investigating}. However, although humans write the expression from left to right in a sequence, the mathematical expression has a structure indeed. Therefore, \citet{xie2019goal} proposes a goal-driven tree-structured model to generate the expression tree during the decoding process. This Seq2Tree approach significantly improved the performance over the traditional Seq2Seq approaches and became the majority approach of neural MWP solvers. Later works improve the Seq2Tree model by leveraging the semantic information and quantity relations~\citep{zhang2020graph} or external knowledge such as syntactic dependency~\citep{lin2021hms} and commonsense knowledge~\citep{wu2020knowledge}. 

Although recent neural-based approaches show some promising results in MWP solving, the SOTA neural-based solver is not able to compete with humans, even on the elementary MWPs. Additionally, it is still unclear whether the neural solvers actually understand the problem and perform the reasoning procedure like humans~\citep{sundaram2022nlp}. \citet{patel2021nlp} discovers that the neural-based approaches can still generate the expression and solution even if we discard the question part of a problem. They suggest that the neural networks may only learn the shallow heuristic to generate the expression rather than perform the actual mathematical reasoning. Additionally, most of the neural solvers do not enforce the mathematical laws (i.e., the commutative, associative, and distributive laws) into the design~\citep{faldu2021towards}, where the representation for the mathematical equivalent expressions are different. Although these equivalent expressions compute the same quantity, the neural solvers would bias toward generating a specific expression, making the prediction not robust. To address this issue, some works focus on generating diverse mathematical equivalent expressions during encoding, where they either apply multiple decoders to generate multiple expressions~\citep{zhang2020teacher,shen2020solving} or enhance the datasets with more golden expressions~\citep{yang2022unbiased}. However, the improvement of these methods is very limited, and the issue has still remained.

\section{Problem Formulation}

A math word problem can be described as a set of words in a natural language containing $l$ words $\mathcal{S} = \{w_1, w_2, ..., w_l\}$. Given the set of quantities $\mathcal{Q} = \{q_1, q_2, ..., q_n\} \subset \mathcal{S}$, a fixed set of constants $\mathcal{C} = \{c_1, c_2, ..., c_m\}$, and a fixed set of binary operators $\mathcal{O} = \{o_1, o_2, ..., o_k\}$, the model aims to generate a $T$-step mathematical expression list which leads to the final answer. The expression list can be formulated as $\mathcal{E} = \{e_1, e_2, ..., e_T\}$, where $e_t = (e_i, e_j, o_t)$ such that $i < t \ \text{or} \ e_i \in \mathcal{Q}$, $j < t \ \text{or} \ e_j \in \mathcal{Q}$, and $o_t \in \mathcal{O}$. $e_t$ denotes the expression of applying the binary operator $o_t$ over quantities or previously generated intermediate mathematical expressions. The final mathematical expression for the problem is $e_T$.

\section{Methodology}
\begin{figure*}[t]
    \centering
    \includegraphics[width=\textwidth]{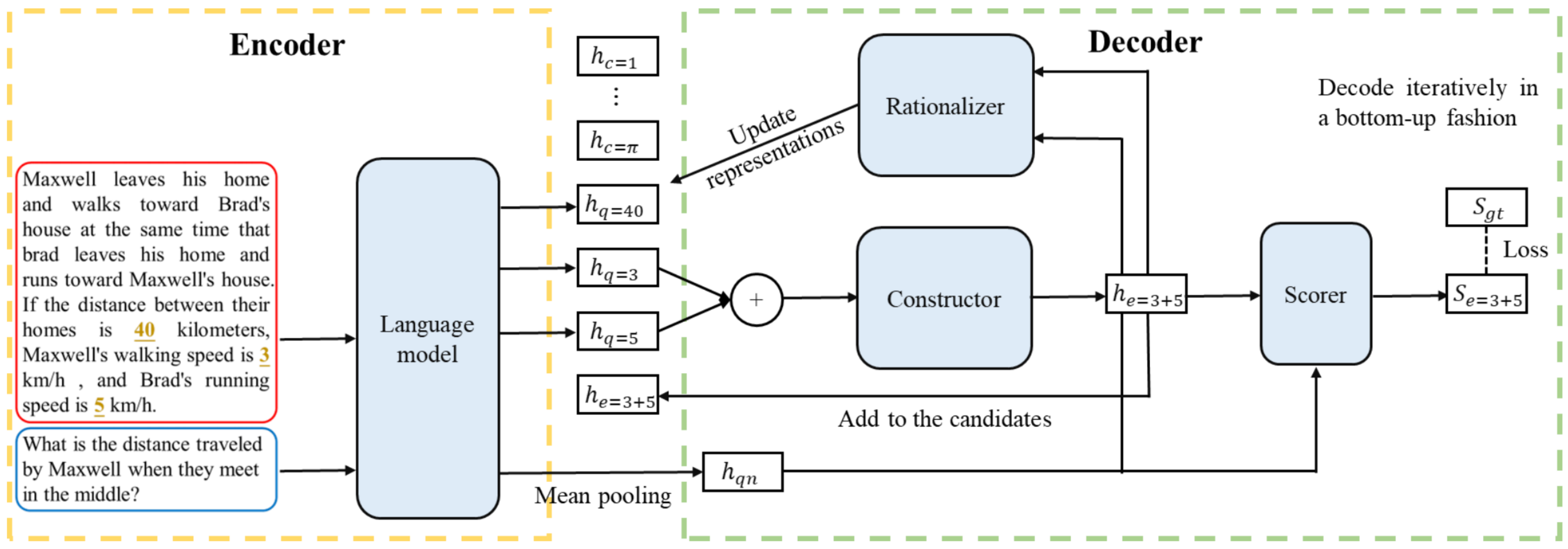}
    \caption{The overview of our framework. Given the quantity embeddings and question embedding from our encoder, our decoder generates the expression $e = 3 + 5$ at the first step.}
    \label{fig:mwp_overview}
\end{figure*}

We propose a novel encoder-decoder framework that fuses the knowledge of question part information into the decoding process and preserves step-wise invariance for mathematical-equivalent expressions. The overview of our model is shown in figure~\ref{fig:mwp_overview}.

Our encoder first uses a language model to get the representation for each quantity in the problem text and the representation for the whole question text. Given both quantity and question embeddings, our decoder follows a bottom-up manner that generates the expression iteratively. At each step, the decoder generates the mathematical expression conditioned on the question embedding, using the question text to guide the decoding process. More details of our encoder and decoder are in the following subsections.

\subsection{Encoder}
A language model is first applied to the raw problem text to acquire the representations of the quantities that appeared in the text. In specific, the quantities in the text are converted into the token ``<\textit{quant}>'' and the converted text is tokenized and fed into the language model. In practice, we use RoBERTa~\citep{liu2019roberta, cui2021pre} as our encoder to generate the representation $h_q$ for each quantity $q$.

Besides generating the representations for the quantities in the text, we also produce the representation for the question text. Note that the question text may not be defined explicitly in some MWPs, so we define the question part $\mathcal{S}_{qn} \subset \mathcal{S}$ as the last complete sentence separated by a full stop of the problem text. To obtain the question representation $h_{qn}$, a mean pooling is applied to all embeddings of the question tokens.

\subsection{Decoder}
Our decoder consists of three components: a constructor, a scorer, and a rationalizer.

\subsubsection{Constructor}

\begin{figure}[t]
    \centering
    \includegraphics[width=\columnwidth]{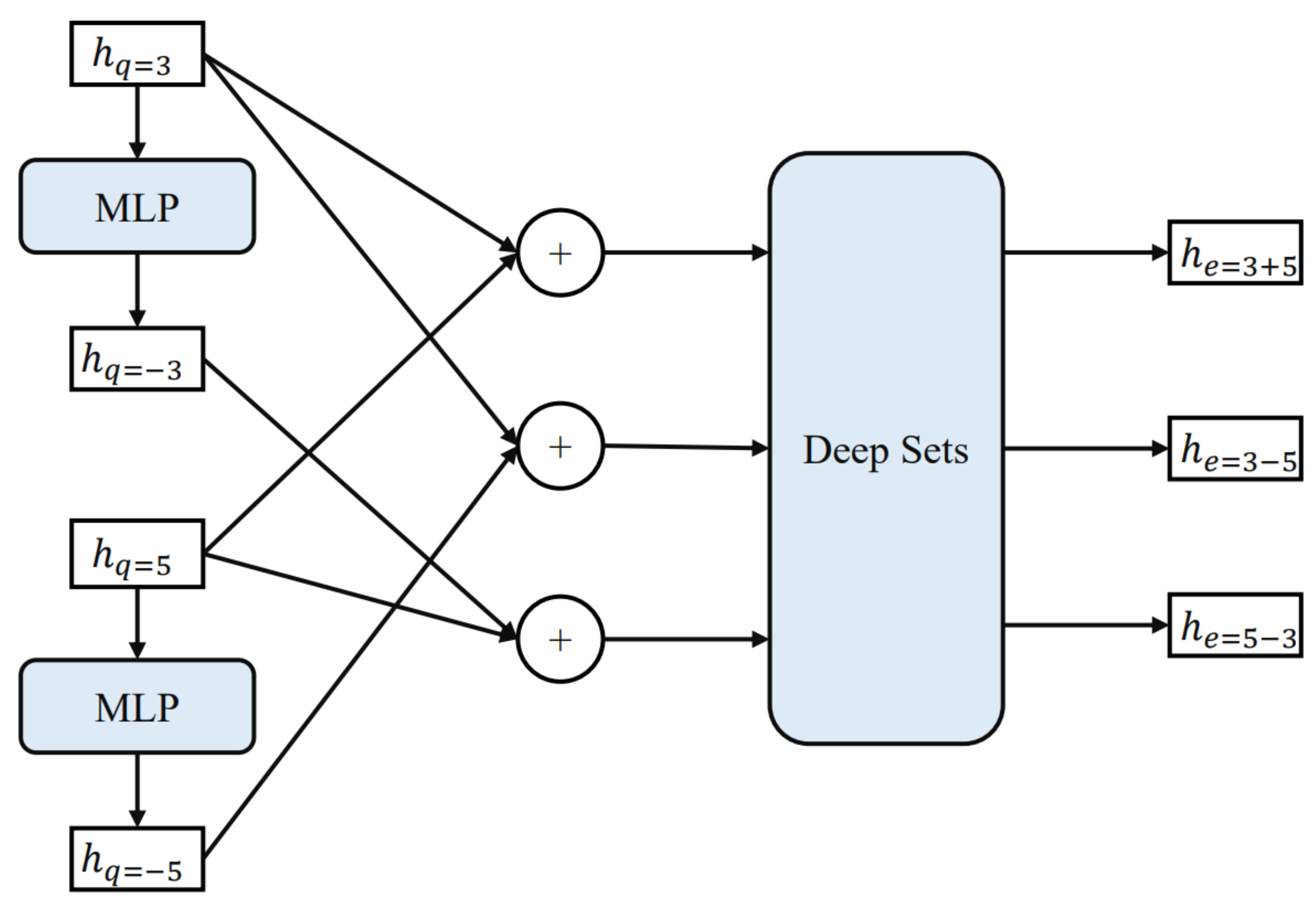}
    \caption{The architecture of our constructor.}
    \label{fig:mwp_constructor}
\end{figure}

Given the embeddings of quantities, our constructor first enumerates all the possible combinations of the quantities and operators at each step. Each combination is a candidate expression for the current step, and we choose one of them as the output expression at this step. To compute the representation for each candidate, the constructor uses a permutation-invariant function for commutative operators like addition and multiplication and a nonlinear transformation for non-communitative operators like additive inverse and multiplicative inverse. In our implementation, we use Deep Sets for communitative binary operators and MLPs for non-communitative unary operators. An illustration of our constructor is shown in figure~\ref{fig:mwp_constructor}. To compute the representation of expression $e = 3 - 5$, we first apply a MLP on the quantity embedding $h_{q=5}$ to obtain the representation $h_{q=-5}$ and uses Deep Sets~\citep{zaheer2017deep} on the two quantity embeddings:
\begin{align}
    h_{q=-5} &= \text{MLP}_{\text{add\_inv}}(h_{q=5})
\end{align}
\begin{align}
    h_{e=3-5} = \text{MLP}_{\text{add}}^{2}\left(\sum_{v = {3,-5}}\text{MLP}_{\text{add}}^{1}(h_{q=v})\right)
\end{align}

It should be stressed that the representations of candidate expressions are invariant under any permutation of the quantities (e.g., $3 - 5$ and $-5 + 3$, $3 \div 5$ and $1 / 5 \times 3$), so our constructor guarantees step-wise commutative law and unifies the representation of expression using $+$ and $-$ operators or $\times$ and $\div$ operators.

\subsubsection{Scorer}
\begin{figure}[t]
    \centering
    \includegraphics[width=\columnwidth]{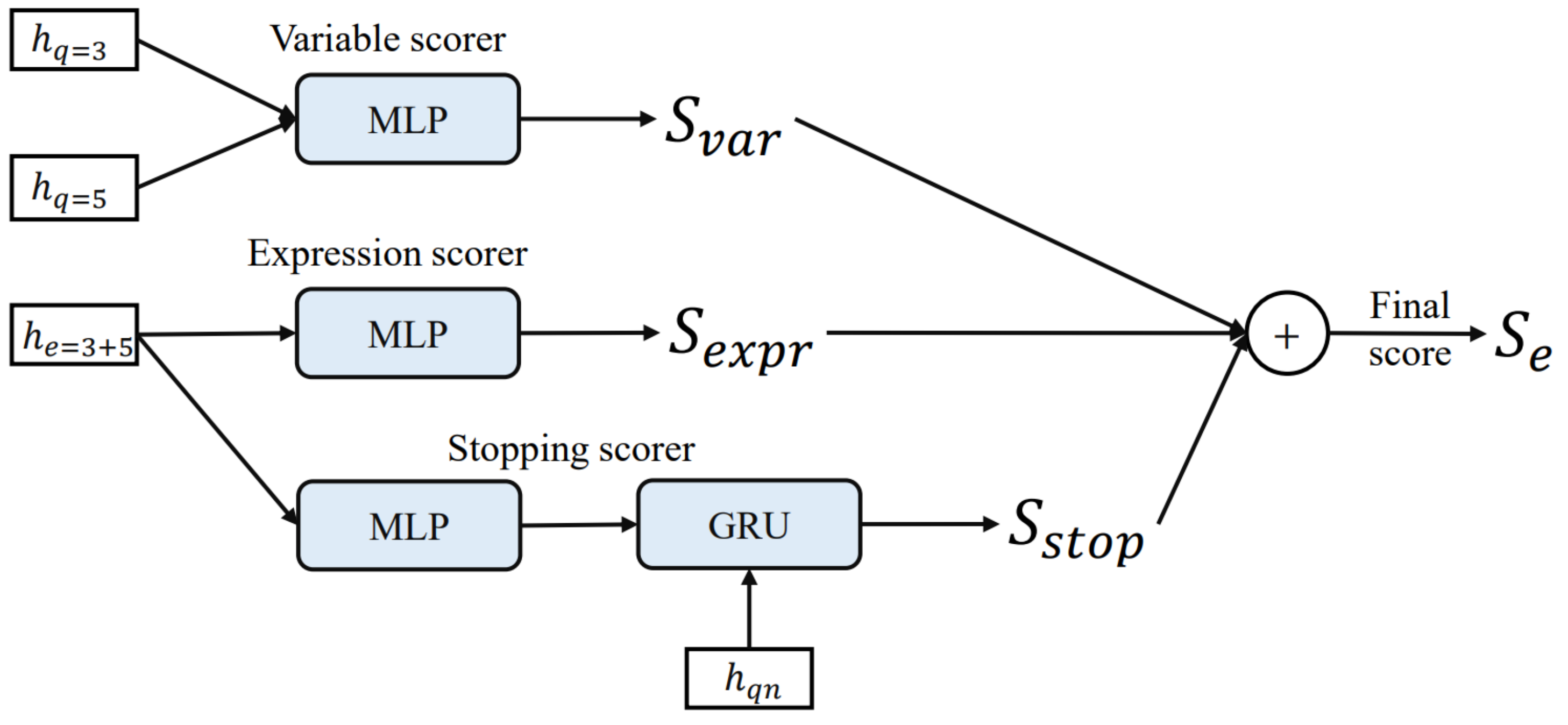}
    \caption{The architecture of our scorer.}
    \label{fig:mwp_scorer}
\end{figure}

With the representations of all candidate expressions generated from our constructor, our scorer scores each of them to decide which expression should be generated at the current step and whether it is the final expression. Figure~\ref{fig:mwp_scorer} shows the architecture of our scorer.

Our scorer consists of three parts: a variable scorer, an expression scorer, and a stopping scorer. The variable scorer and expression scorer compute a continuation score, which refers to the score of choosing the expression as an intermediate expression. The stopper computes the termination score, indicating the score of choosing the expression as a final result expression. Specifically, the variable scorer takes the representations of the two quantities of the expression and computes the variable score $s_{\text{var}}$, representing the probability that quantities $q_i, q_j$ should be used at the current step:
\begin{align}
    s_{\text{var}} &= \text{MLP}_{\text{var}}(h_{q = q_i}) + \text{MLP}_{\text{var}}(h_{q = q_j})
\end{align}

The expression scorer takes the representation of the candidate expression $h_e$ and computes the expression score $s_{\text{expr}}$ using a MLP:
\begin{align}
    s_{\text{expr}} &= \text{MLP}_{\text{epxr}}(h_e)
\end{align}

To determine whether the decoding process should be stopped, the stopping scorer takes each representation of candidate expression $h_e$ and question $h_{qn}$ as input and computes the stopping score $s_{\text{stop}}$ using a MLP and a GRU cell~\citep{chung2014empirical}:
\begin{align}
    s_{\text{stop}} &= \text{GRU}_{\text{stop}}\left(\text{MLP}_{\text{stop}}(h_e), h_{qn}\right)
\end{align}

Finally, the score for the candidate expression $s_e$ can be obtained by the sum of the three parts:
\begin{align}
    s_{e} &= s_{\text{var}} + s_{\text{expr}} + s_{\text{stop}}
\end{align}

\subsubsection{Rationalizer}
Following \citet{jie2022learning}, we also adapt a rationalizer to update the representations of all quantities and intermediate expressions at the end of each step. This module is crucial because if the representations are not updated, those expressions that were initially highly ranked would always be preferred. For example, the expression $e = 3 + 5$ is scored the highest in the first step, it is likely scored very high in the latter step if its representation remains the same.

\begin{figure}[t]
    \centering
    \includegraphics[width=\columnwidth]{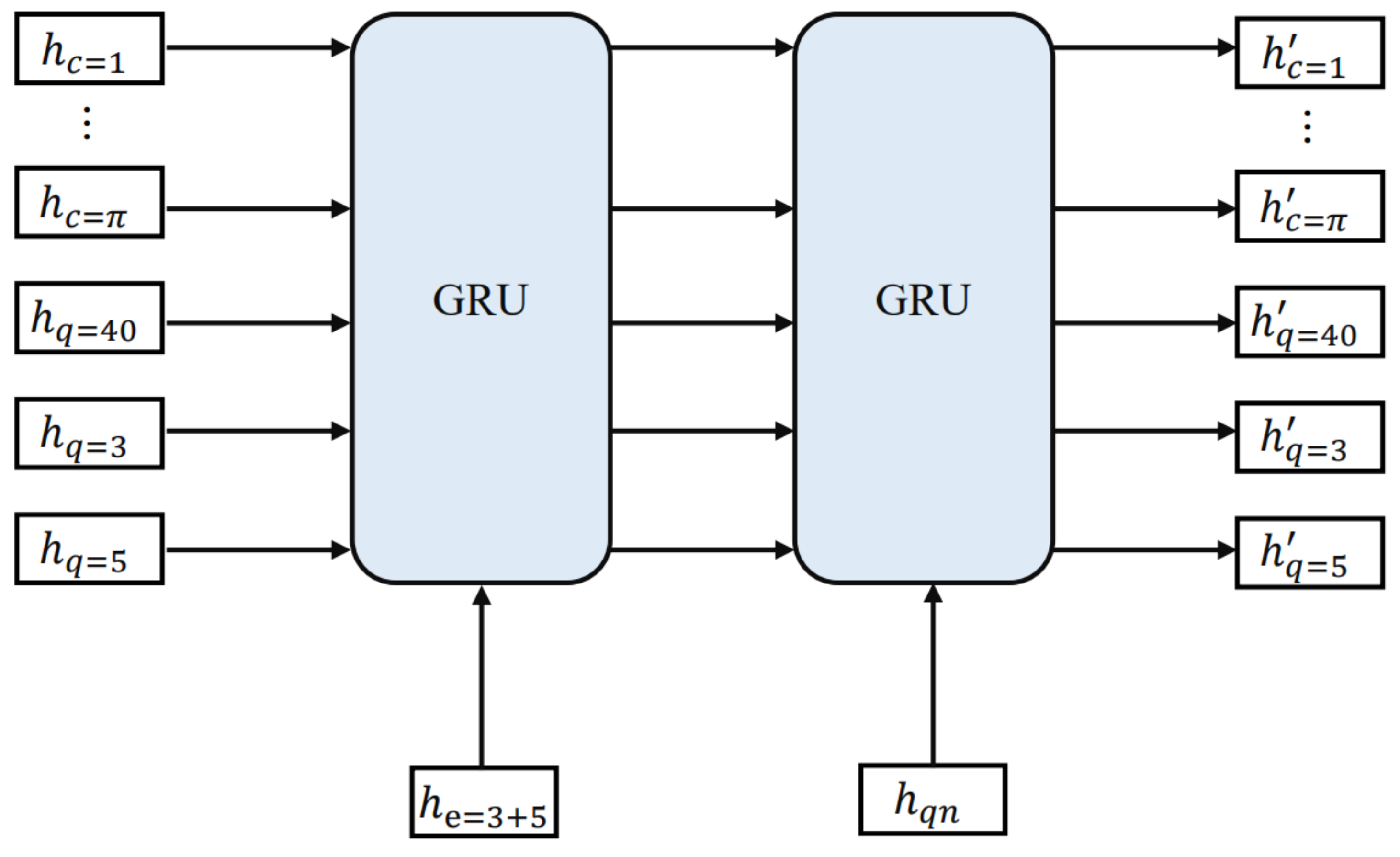}
    \caption{Architecture of rationalizer.}
    \label{fig:mwp_rationalizer}
\end{figure}

Figure~\ref{fig:mwp_rationalizer} shows the architecture of our rationalizer. In our implementation, we rationalize the quantity representations using the newly chosen expression and question representations using two GRU cells:
\begin{align}
    h^{'}_{q} &= \text{GRU}_{\text{rat}}^{2}\left(h_{qn}, (\text{GRU}_{\text{rat}}^{1}(h_q, h_e)\right)
\end{align}

The first GRU cell takes the quantity representations as input and the new expression representation as the hidden state. The second GRU cell takes the representations from the first GRU cell as input and the question representation as the hidden state. The final output is the updated representation for each quantity.

\subsection{Training and Inference}
We adopt the teacher-forcing strategy~\citep{williams1989learning} to guide the model with the golden expression at each step during training. Denote all the learnable parameters in our framework as $\theta$, our loss is defined as:
\begin{align}
    \mathcal{L}_\theta = \sum_{t=1}^T \left(\max_{e \in \text{candidates}^t} s_e - s_{e_t^*}\right)
\end{align}
where $T$ is the total step to generate the ground truth expression, $\text{candidates}^t$ represents all the candidate expressions at step $t$, and $s_{e_t^*}$ is the ground truth expression at step $t$. For each step, this loss minimizes the gap between the scores of the gold expression and the expression with the highest score, encouraging our framework to generate the highest score for the ground truth expression.

During inference, we iteratively choose the expression with the highest score at each step until $t$ reaches a predefined maximum step $T$. For the generated expressions $e_{1}, e_{2}, \cdots, e_{T}$, the expression with the highest termination score is chosen as the output expression and its corresponding numerical answer is computed as the final answer.

\section{Experiments}

\subsection{Datasets}
We conduct experiments on four established MWP benchmarks: MAWPS~\citep{koncel2016mawps}, Math23k~\citep{wang2017deep}, SVAMPS~\citep{patel2021nlp}, and MathQA~\citep{amini2019mathqa}. Table~\ref{tab:dataset_stats} shows the statistics of these datasets. MAWPS and Math23k are commonly used in previous research and contain primary school math questions. MathQA and SVAMP are much more challenging. MathQA contains more complex GRE questions in many domains including physics, geometry, and probability, and therefore has a large portion of equations with more operations. SVAMP contains manually designed challenging questions created by applying variations over the problems from MAWPS, which has lots of unrelated quantities and context text. See Appendix~\ref{app:samples} for problem samples from these four datasets.

\begin{table}[t]
\resizebox{\columnwidth}{!}{
\begin{tabular}{lccccc}
\hline
Dataset & \#Train & \#Valid & \#Test & Operations & Language \\
\hline
MAWPS   & 1,589   & 199     & 199    & + - $\times$ $\div$          & English  \\
Math23k & 21,162  & 1,000   & 1,000  & + - $\times$ $\div$ \^           & Chinese  \\
MathQA  & 16,191  & 2,411   & 1,605  & + - $\times$ $\div$ \^          & English  \\
SVAMP   & 3,138   & -       & 1,000  & + - $\times$ $\div$           & English  \\ 
\hline
\end{tabular}
}
\caption{Dataset statistics.}
\label{tab:dataset_stats}
\end{table}

\subsection{Evaluation Metric}
We use the final value accuracy (Val Acc.) as our evaluation metric, indicating the accuracy for which our computed numerical answer from the equation equals the ground-truth answer. Note that there can be multiple mathematical expressions that lead to the same numerical answer (e.g., $(1 + 2) \times 3$ and $1 \times 3 + 2 \times 3$), so generating any of them is considered correct.

\subsection{Baselines}
The baseline approaches are categorized into sequence-to-sequence (Seq2Seq) models and sequence-to-tree (Seq2Tree) models. We compare our proposed model against four Seq2Seq models and six Seq2Tree models on the four benchmarks. Among those Seq2Seq MWP solvers, \textbf{GroupAttn}~\citep{li2019modeling} proposes to design several types of attention mechanisms such
as quantity-related attentions in the seq2seq model. \textbf{mBERT-LSTM}~\citep{tan2021investigating} uses multilingual BERT as the encoder and LSTM as the decoder. \textbf{BERT-BERT}~\citep{lan2022mwptoolkit} and \textbf{RoBERTa-RoBERTa}~\citep{lan2022mwptoolkit} employs BERT and RoBERTa as both encoder and decoder. Among those Seq2Tree models, \textbf{GTS}~\citep{xie2019goal} is the seminal work that uses the bidirectional GRU to encode the problem text and decode the mathematical expression using a tree structure in a top-down manner. \textbf{Graph2Tree}~\citep{zhang2020graph} enhances the encoder of GTS by modeling the quantity relationships and order information using a GCN. \textbf{BERT-Tree}~\citep{liang2021mwp} pre-trains BERT using 8 pre-training tasks to solve the number representation issue and uses it as the encoder. \textbf{RoBERTa-GTS} and \textbf{RoBERTa-Graph2Tree} replace the original encoder of GTS and Graph2Tree with RoBERTa. The most similar work of our approach is \textbf{RoBERTa-DeductiveReasoner}~\citep{jie2022learning}, which uses RoBERTa as the encoder and decodes the mathematical expression using a bottom-up approach. However, it encodes the problem text as a whole without any special attention to the question text and fails to preserve any mathematical law when computing the representation for each candidate expression.

\subsection{Implementation Details}
For the English datasets MAWPS, MathQA, and SvAMP, we use RoBERTa~\citep{liu2019roberta} as the encoder. For the Chinese dataset Math23K, we use Chinese RoBERTa~\citep{cui2021pre} as the encoder. The pre-trained RoBERTa models are initialized from HuggingFaces Transformers~\citep{wolf2020transformers}. All MLPs have 2 hidden layers with 768 units each and use ReLU as the activation function. We use a batch size of $30$ when training on the 4 datasets. On Math23k, MathQA, and SVAMP, we train our model for $1,000$ epochs. On MAWPS, we train the model for $100$ epochs. The Adam optimizer with a learning rate of $2\times 10^{-5}$ is used to optimize our loss function. All experiments are run 3 times for each dataset on a cluster with 4 NVIDIA RTX-8000 GPUs. Following previous works~\citep{lan2022mwptoolkit, jie2022learning}, we report the 5-fold cross-validation accuracy on MAWPS and the test accuracy on Math23k, MathQA, and SVAMP. The results of the baseline approaches are taken from their original papers.

\subsection{Main Results}

\begin{table}[t]
\resizebox{\columnwidth}{!}{
\begin{tabular}{llc}
\hline
\multicolumn{1}{c}{\textbf{Class}} & \multicolumn{1}{c}{\textbf{Model}} & \textbf{Val Acc.} \\ \hline
\multirow{4}{*}{Seq2Seq}  & GroupAttn                 & 76.1     \\
                          & Transformer               & 85.6     \\
                          & BERT-BERT                 & 86.9     \\
                          & RoBERTa-RoBERTa           & 88.4     \\ \hline
\multirow{5}{*}{Seq2Tree} & GTS                       & 82.6     \\
                          & Graph2Tree                & 85.6     \\
                          & RoBERTa-GTS               & 88.5     \\
                          & RoBERTa-Graph2Tree        & 88.7     \\
                          & RoBERTa-DeductReasoner    & 92.0 $\pm$ 0.20     \\ \hline
                          & \textbf{Ours}                      & \textbf{92.9 $\pm$ 0.14}     \\ \hline
\end{tabular}
}
\caption{5-fold cross-validation result comparison on MAWPS.}
\label{tab:result_mawps}
\end{table}

\begin{table}[t]
\resizebox{\columnwidth}{!}{
\begin{tabular}{llc}
\hline
\multicolumn{1}{c}{\textbf{Class}} & \multicolumn{1}{c}{\textbf{Model}} & \textbf{Val Acc.} \\ \hline
\multirow{2}{*}{Seq2Seq}                   & GroupAttn                 & 69.5     \\
                          & mBERT-LSTM                & 75.1     \\ \hline
\multirow{3}{*}{Seq2Tree}                  & GTS                       & 75.6     \\
                          & Graph2Tree                & 77.4     \\
                          & RoBERTa-DeductReasoner    & 85.1 $\pm$ 0.24     \\ \hline
                          & \textbf{Ours}                      & \textbf{85.3 $\pm$ 0.21}     \\ \hline
\end{tabular}
}
\caption{Test accuracy comparison on Math23k.}
\label{tab:result_math23k}
\end{table}

\begin{table}[t]
\resizebox{\columnwidth}{!}{
\begin{tabular}{llc}
\hline
\multicolumn{1}{c}{\textbf{Class}} & \multicolumn{1}{c}{\textbf{Model}} & \textbf{Val Acc.} \\ \hline
Seq2Seq                   & BERT-LSTM                 & 77.1     \\ \hline
\multirow{3}{*}{Seq2Tree}                  & Graph2Tree                & 69.5     \\
                          & BERT-Tree                 & 73.8     \\
                          & RoBERTa-DeductReasoner    & 77.2 $\pm$ 0.11     \\ \hline
                          & \textbf{Ours}                      & \textbf{81.1 $\pm$ 0.13}     \\ \hline
\end{tabular}
}
\caption{Test accuracy comparison on MathQA.}
\label{tab:result_mathqa}
\end{table}

\begin{table}[t]
\resizebox{\columnwidth}{!}{
\begin{tabular}{llc}
\hline
\multicolumn{1}{c}{\textbf{Class}} & \multicolumn{1}{c}{\textbf{Model}} & \textbf{Val Acc.} \\ \hline
\multirow{3}{*}{Seq2Seq}  & GroupAttn                 & 21.5     \\
                          & BERT-BERT                 & 24.8     \\
                          & RoBERTa-RoBERTa           & 30.3     \\ \hline
\multirow{3}{*}{Seq2Tree}                  & GTS                       & 30.8     \\
                          & Graph2Tree                & 36.5     \\
                          & RoBERTa-DeductReasoner    & 44.0 $\pm$ 0.20     \\ \hline
                          & \textbf{Ours}                      & \textbf{44.4 $\pm$ 0.11}     \\ \hline
\end{tabular}
}
\caption{Test accuracy comparison on SVAMP.}
\label{tab:result_svamp}
\end{table}

Table~\ref{tab:result_mawps} shows the 5-fold cross-validation results for different models on the MAWPS dataset. Table~\ref{tab:result_math23k}, table~\ref{tab:result_mathqa}, table~\ref{tab:result_svamp} show the test accuracy comparison on the Math23k, MathQA, and SVAMP dataset respectively. The results show that Seq2Tree models generally have better performances compared with Seq2Seq models, where we conjecture this is because the tree decoder can better handle the structure of a mathematical expression than the sequence decoder. Meanwhile, the performances of the models like GTS and Graph2Tree are improved significantly by using pre-trained large language models like BERT or RoBERTa as the encoder, indicating the representation ability of large language models is useful in MWP solving.

The test accuracies of our model on MAWPS, Math23k, MathQA, and SVAMP are $92.9\%$, $85.3\%$, $81.1\%$, and $44.4\%$ respectively. On all the benchmarks, our model achieves state-of-the-art performances, which surpasses the best baseline model RoBERTa-DeductReasoner by $0.9\%$, $0.2\%$, $3.9\%$, and $0.4\%$ respectively. Most notably, our model improves the vanilla RoBERTa-DeductReasoner by $3.9\%$ on the MathQA dataset, which is the largest and hardest dataset of the four. These significant improvements show that both leveraging the question text information and enforcing mathematical laws into the design are useful in MWP solving. Conclusively, the comparisons well demonstrate the effectiveness of our proposed techniques to improve neural MWP solvers.

\subsection{Ablation Study}

To analyze our techniques in detail, we further perform an ablation study to explore the effectiveness of our proposed two techniques. We replace our constructor with the module in RoBERTa-DeductReasoner~\citep{jie2022learning} as the model without preserving communitative law and discard the question representation in our decoder as the model without question representation. The results are shown in Table~\ref{tab:ablation_study}.
Compared with the model without preserving the step-wise communitative law, our complete model improves the test accuracies by $0.3\%$, $1.2\%$, $0.2\%$ on MAWPS, MathQA, and SVAMP respectively. Exceptionally, the accuracy decreases by $0.5\%$ on Math23k. We conjecture this is because the mathematical expression in MathQA is relatively complex so only preserving step-wise communitative law for the expressions is not good enough. On the other hand, the complete model improves the test accuracies of the one without question embedding by $0.6\%$, $0.7\%$ $2.9\%$, $0.6\%$ on MAWPS, Math23k, MathQA, and SVAMP. These results demonstrate that both enforcing mathematical law and explicitly modeling the question text to guide the decoding procedure with the question representation can generally improve neural MWP solvers. The performance of the model can be further improved by applying both techniques.  By comparing these two strategies, we can observe that the exploitation of question part information provides greater improvement to the accuracy than preserving step-wise communitative law for mathematical expressions.

\begin{table}[t]
\resizebox{\columnwidth}{!}{
\begin{tabular}{llc}
\hline
\multicolumn{1}{c}{\textbf{Dataset}} & \multicolumn{1}{c}{\textbf{Model}}        & \textbf{Val Acc.} \\ \hline
\multirow{4}{*}{MAWPS}      & RoBERTa-DeductReasoner           & 92.0     \\
& \textbf{Ours}             & \textbf{92.9} \\
                            &  - w/o preserving communitative law     & 92.6     \\
                            &  - w/o question representation & 92.3     \\\hline
\multirow{4}{*}{Math23k}    & RoBERTa-DeductReasoner           & 85.1     \\
& \textbf{Ours}              & 85.3     \\ 
                            & - w/o preserving communitative law     & \textbf{85.8}     \\
                            & - w/o question representation & 84.6     \\
                            \hline
\multirow{4}{*}{MathQA}     & RoBERTa-DeductReasoner           & 77.2     \\
& \textbf{Ours}              & \textbf{81.0}     \\ 
                            & - w/o preserving communitative law     & 79.8     \\
                            & - w/o question representation & 78.1     \\
                            \hline
\multirow{4}{*}{SVAMP}      & RoBERTa-DeductReasoner           & 44.0     \\
& \textbf{Ours}              & \textbf{44.4}     \\ 
                            & - w/o preserving communitative law     & 44.2     \\
                            & - w/o question representation & 43.8     \\
                            \hline
\end{tabular}
}
\caption{Ablation study of our model. Our model is compared with the one without question part information and the one without preserving communitative law.}
\label{tab:ablation_study}
\end{table}

\section{Discussion}

\subsection{Limitation}
Our current model provides a technique to enforce the commutative law and unify representations for a group of operators like addition and subtraction or multiplication and division at each step. For example, the representations $h_{e = 1 + 2} = h_{e = 2 + 1}$ and $h_{e = 2 - 1} = h_{e = 2 + (-1)} = h_{e = (-1) + 2}$. However, the commutative law is not preserved when computing a complex expression that contains more than a single operator. For example, $h_{e = (1 + 2) + 3} \ne h_{e = (3 + 2) + 1}$, where different orders of the operation result in a different representation. Moreover, other mathematic laws like distributive law and associative law also fail to be preserved. For example, $h_{e = (1 + 2) \times 3} \ne h_{e = 1 \times 3 + 2 \times 3}$.

\subsection{Future Work}
In the future, we plan to enforce more mathematical laws in neural MWP solvers and build an invariant representation for all mathematically equivalent expressions. This idea could be implemented by designing a unified format to represent all mathematical expressions and compute the representations based on such a form. Specifically, we could represent all the expressions in a fine-grained format with no parentheses. For example, to compute the representation of $(1 + 2) \times 3$, instead of generating the embeddings for (1 + 2) first and then $(1 + 2) \times 3$, we could first convert the expression to $1 \times 3 + 2 \times 3$ and then compute the representation by applying some permutation invariant functions on such a form. Therefore, all mathematical expressions would have the same embeddings and the model can also preserve the distributive law and the associative law.

\subsection{Conclusion}
Existing neural math word solvers mostly fail to fully leverage the question text or preserve any mathematical laws. In this work, we propose a new encoder-decoder framework that applies two techniques to address these issues: (1) our encoder generates an embedding for the question text and uses it to guide the decoding process. (2) our decoder applies Deep Sets to compute the representations of candidate expressions to enforce step-wise communitative law. Experiments on four standard MWP benchmarks show that these two techniques could improve the performance of neural MWP solvers and make our model achieve state-of-the-art performance. 


\subsection{Acknowledgement}
I would like to thank Mr. Zhaoyu Li, Prof. Xujie Si and Siva Reddy for their assistance in this work. Special thanks to Mr. Zhaoyu Li who surveyed related publications and proposed the idea.

\bibliography{anthology,custom}
\bibliographystyle{acl_natbib}

\appendix
\section{Appendix}

\subsection{Samples from Datasets}
\label{app:samples}
Table~\ref{tab:dataset_samples} shows two samples for each dataset.

\begin{CJK*}{UTF8}{gbsn}

\begin{table*}[ht]
\begin{tabularx}{\textwidth}{lX}
\hline

Dataset    & \multicolumn{1}{c}{MAWP}                                                                                                                          \\ \hline
Input      & Mary is baking a cake. The recipe wants 8 cups of flour. She already put in 2 cups. How many cups does she need to add?                           \\
Answer     & 6                                                                                                                                                 \\
Expression & 8 - 2                                                                                                                                             \\ \hline
Input      & There are 139 erasers and 118 scissors in the drawer. Jason placed 131 erasers in the drawer. How many erasers are now there in total?            \\
Answer     & 270                                                                                                                                               \\
Expression & 139 + 131                                                                                                                                         \\ \hline
\\ \hline
Dataset    & \multicolumn{1}{c}{Math23k}                                                                                                                       \\ \hline
Input      & 一个工程队挖土，第一天挖了316方，从第二天开始每天都挖230方，连续挖了6天，这个工程队一周共挖土多少方？                                                                                            \\
Translation& An engineering team dug 316 cubic meters on the first day and 230 cubic meters every day from the second day, which lasted for 6 consecutive days. How many cubic meters did the team dig in a week?                                                                                            \\
Answer     & 1466                                                                                                                                              \\
Expression & 316 + 230 $\times$ (6 - 1)                                                                                                                               \\ \hline
Input      & 从甲地到乙地，如果骑自行车每小时行驶16千米，4小时可以到达，如果乘汽车只需要2小时，汽车每小时行驶多少千米？                                                                                           \\
Translation&From place A to place B, if you travel 16 kilometers an hour by bike, you can get there in 4 hours. If it takes only 2 hours by car, how many kilometers does the car travel per hour? \\
Answer     & 32                                                                                                                                                \\
Expression & 16 $\times$ 4 $\div$ 2                                                                                                                                        \\ \hline
\\ \hline
Dataset    & \multicolumn{1}{c}{MathQA}                                                                                                                        \\ \hline
Input      & Sophia finished 2/3 of a book. She calculated that she finished 90 more pages than she has yet to read. How long is her book?                   \\
Answer     & 270                                                                                                                                               \\
Expression & 90 $\div$ (1 - 2 $\div$ 3)                                                                                                                                  \\ \hline
Input      & 6 workers should finish a job in 8 days. After 3 days came 4 workers join them. How many days do they need to finish the same job?                \\
Answer     & 2.5                                                                                                                                               \\
Expression & (6 $\times$ 8 - (6 $\times$ 3)) $\div$ (8 + 4)                                                                                                                       \\ \hline
\\ \hline
Dataset    & \multicolumn{1}{c}{SVAMP}                                                                                                                         \\ \hline
Input      & Bobby ate some pieces of candy. then he ate 25 more. If he ate a total of 43 pieces of candy. How many pieces of candy had he eaten at the start? \\
Answer     & 18                                                                                                                                                \\
Expression & 43 - 25                                                                                                                                           \\ \hline
Input      & Bobby ate 28 pieces of candy. Then he ate 42 more. He also ate 63 pieces of chocolate. How many pieces of candy did bobby eat?                    \\
Answer     & 70                                                                                                                                                \\
Expression & 28 + 42                                                                                                                                           \\ \hline
\end{tabularx}
\caption{Samples from MAWP, Math23k, MathQA, and SVAMP.}
\label{tab:dataset_samples}
\end{table*}
\end{CJK*}

\end{document}